\newcommand{\omt}[1]{}
\newcommand{\SC}{\scriptstyle}
\newcommand{\RHnorm}[1]{\tilde #1}

\documentclass[11pt]{article}
\usepackage{colacl, psfig}

\title{An Empirical Evaluation of Probabilistic Lexicalized Tree
Insertion Grammars}

\author{Rebecca Hwa                                    \\
     Harvard  University                                    \\ 
	 Cambridge, MA 02138 USA                           \\
	 rebecca@eecs.harvard.edu
}

\begin{document}
\maketitle
\begin{abstract}
We present an empirical study of the applicability of Probabilistic
Lexicalized Tree Insertion Grammars (PLTIG), a lexicalized counterpart
to Probabilistic Context-Free Grammars (PCFG), to problems in
stochastic natural-language processing.  Comparing the performance of
PLTIGs with non-hierarchical $N$-gram models and PCFGs, we show that
PLTIG combines the best aspects of both, with language modeling capability
comparable to $N$-grams, and improved parsing performance over its
non-lexicalized counterpart. Furthermore, training of PLTIGs displays
faster convergence than PCFGs.
\end{abstract}

\bibliographystyle{acl}

\section{Introduction}

There are many advantages to expressing a grammar in a {\it lexicalized
form}, where an observable word of the language is encoded in each
grammar rule.  First, the lexical words help to clarify ambiguities that
cannot be resolved by the sentence structures alone.  For example, to
correctly attach a prepositional phrase, it is often necessary to
consider the lexical relationships between the head word of the
prepositional phrase and those of the phrases it might modify.  Second,
lexicalizing the grammar rules increases computational efficiency because
those rules that do not contain any observed words can be pruned away
immediately.  The Lexicalized Tree Insertion Grammar formalism (LTIG) has
been proposed as a way to lexicalize context-free grammars
\cite{Schabes:94a}.  We now apply a probabilistic variant of this
formalism, Probabilistic Tree Insertion Grammars (PLTIGs), to natural
language processing problems of stochastic parsing and language modeling.
This paper presents two sets of experiments, comparing PLTIGs with
non-lexicalized Probabilistic Context-Free Grammars (PCFGs)
\cite{Pereira:92a} and non-hierarchical $N$-gram models that use the
right branching bracketing heuristics (period attaches high) as their
parsing strategy.  We show that PLTIGs can be induced from partially
bracketed data, and that the resulting trained grammars can parse unseen
sentences and estimate the likelihood of their occurrences in the
language.  The experiments are run on two corpora: the Air Travel
Information System (ATIS) corpus and a subset of the Wall Street Journal
TreeBank corpus.  The results show that the lexicalized nature of the
formalism helps our induced PLTIGs to converge faster and provide a
better language model than PCFGs while maintaining comparable parsing
qualities.  Although $N$-gram models still slightly out-perform PLTIGs on
language modeling, they lack high level structures needed for parsing.
Therefore, PLTIGs have combined the best of two worlds: the language
modeling capability of $N$-grams and the parse quality of context-free
grammars.

The rest of the paper is organized as follows: first, we present an
overview of the PLTIG formalism; then we describe the experimental
setup; next, we interpret and discuss the results of the experiments;
finally, we outline future directions of the research.

\section{PLTIG and Related Work}

The inspiration for the PLTIG formalism stems from the desire to
lexicalize a context-free grammar.  There are three ways in which one
might do so.  First, one can modify the tree structures so that all
context-free productions contain lexical items.  Greibach normal form
provides a well-known example of such a lexicalized context-free
formalism.  This method is not practical because altering the
structures of the grammar damages the linguistic information stored in
the original grammar \cite{Schabes:94a}.  Second, one might propagate
lexical information upward through the productions.  Examples of
formalisms using this approach include the work of
\newcite{Magerman:95a},
\newcite{Charniak:97a},
\newcite{Collins:97a}, and \newcite{Goodman:97a}.  A more
linguistically motivated approach is to expand the domain of
productions downward to incorporate more tree structures.  The
Lexicalized Tree-Adjoining Grammar (LTAG) formalism
\cite{Schabes:88a}, \cite{Schabes:90} , although not
context-free, is the most well-known instance in this category.
PLTIGs belong to this third category and generate only 
context-free languages.

LTAGs (and LTIGs) are tree-rewriting systems, consisting of a set of
elementary trees combined by tree operations.  We distinguish two
types of trees in the set of elementary trees: the {\em initial trees}
and the {\em auxiliary trees}.  Unlike full parse trees but
reminiscent of the productions of a context-free grammar, both types
of trees may have nonterminal leaf nodes.  Auxiliary trees have, in
addition, a distinguished nonterminal leaf node, labeled with the same
nonterminal as the root node of the tree, called the {\em foot} node.
Two types of operations are used to construct {\em derived trees}, or
parse trees: substitution and adjunction.  An initial tree can be
{\em substituted} into the nonterminal leaf node of another tree in a
way similar to the substitution of nonterminals in the production
rules of CFGs. An auxiliary tree is inserted into another tree through
the adjunction operation, which splices the auxiliary tree into the
target tree at a node labeled with the same nonterminal as the root
and foot of the auxiliary tree.  By using a tree representation, LTAGs
extend the domain of locality of a grammatical primitive, so that they
capture both lexical features and hierarchical structure.  Moreover,
the adjunction operation elegantly models intuitive linguistic
concepts such as long distance dependencies between words.  Unlike the
$N$-gram model, which only offers dependencies between neighboring
words, these trees can model the interaction of structurally related
words that occur far apart.

Like LTAGs, LTIGs are tree-rewriting systems, but they differ from
 LTAGs in their generative power.  LTAGs can generate some strictly
 context-sensitive languages.  They do so by using {\it wrapping}
 auxiliary trees, which allow non-empty frontier nodes (i.e., leaf
 nodes whose labels are not the empty terminal symbol) on both sides
 of the foot node.  A wrapping auxiliary tree makes the formalism
 context-sensitive because it coordinates the string to the left of
 its foot with the string to the right of its foot while allowing a
 third string to be inserted into the foot.  Just as the ability to
 recursively center-embed moves the required parsing time from $O(n)$
 for regular grammars to $O(n^3)$ for context-free grammars, so the
 ability to wrap auxiliary trees moves the required parsing time
 further, to $O(n^6)$ for tree-adjoining grammars
\footnote{The best theoretical upper bound on time complexity for the
recognition of Tree Adjoining Languages is $O(M(n^2))$, where $M(k)$
is the time needed to multiply two $k \times k$ boolean
matrices.\cite{Rajasekaran:95} }.   This
 level of complexity is far too computationally expensive for current
 technologies.  The complexity of LTAGs can be moderated by
 eliminating just the wrapping auxiliary trees.  LTIGs prevent
 wrapping by restricting auxiliary tree structures to be in one of
 two forms: the {\it left auxiliary tree}, whose non-empty frontier
 nodes are all to the left of the foot node; or the {\it right
 auxiliary tree}, whose non-empty frontier nodes are all to the right
 of the foot node.  Auxiliary trees of different types cannot adjoin
 into each other if the adjunction would result in a wrapping
 auxiliary tree.  The resulting system is strongly equivalent to CFGs,
 yet is fully lexicalized and still $O(n^3)$ parsable, as shown by
 \newcite{Schabes:94a}.

Furthermore, LTIGs can be parameterized to form probabilistic models
\cite{Schabes:93b}.  Appendix \ref{app:param} describes the parameters
in detail.  Informally speaking, a parameter is associated with each
possible adjunction or substitution operation between a tree and a
node.  For instance, suppose there are $V$ left auxiliary trees that
might adjoin into node $\eta$.  Then there are $V+1$ parameters
associated with node $\eta$ that describe the distribution of the
likelihood of any left auxiliary tree adjoining into node $\eta$.  (We
need one extra parameter for the case of no left adjunction.)  A
similar set of parameters is constructed for the right adjunction and
substitution distributions.

\section{Experiments}

In the following experiments we show that PLTIGs of varying sizes and
configurations can be induced by processing a large 
training corpus, and that the trained PLTIGs can provide parses on
unseen test data of comparable quality to the parses produced by
PCFGs.  Moreover, we show that PLTIGs have significantly lower
entropy values than PCFGs, suggesting that they make better language models.
 We describe the induction process of the PLTIGs in Section
\ref{setup}.  Two corpora of very different nature are used for
training and testing.  The first set of experiments uses the Air
Travel Information System (ATIS) corpus.  Section \ref{atis} presents
the complete results of this set of experiments.  To determine if
PLTIGs can scale up well, we have also begun another study that uses a
larger and more complex corpus, the Wall Street Journal TreeBank
corpus.  The initial results are discussed in Section
\ref{wsj}.  To reduce the effect of the data sparsity problem, we back
off from lexical words to using the part of speech tags as the
anchoring lexical items in all the experiments.  Moreover, we use the
deleted-interpolation smoothing technique for the $N$-gram models and
PLTIGs.  PCFGs do not require smoothing in these experiments.

\subsection{Grammar Induction}
\label{setup}

\begin{figure}[t]
\psfig{figure=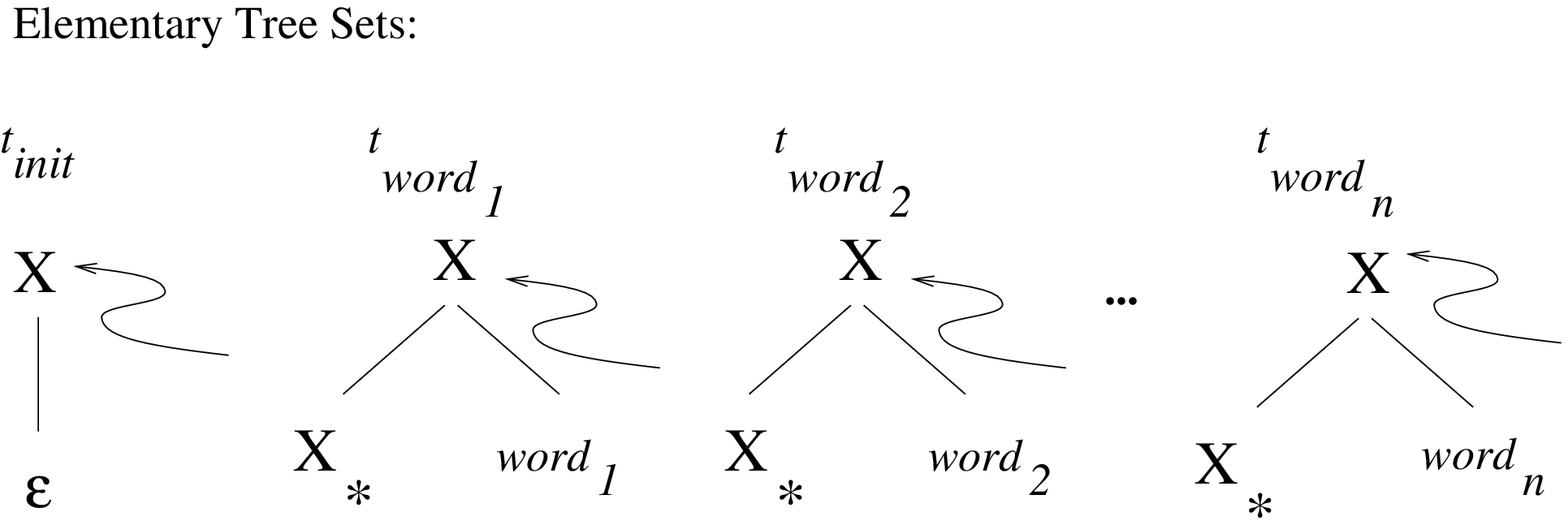,width=2.7in}
\caption{A set of elementary LTIG trees that represent a bigram
grammar.  The arrows indicate adjunction sites.}
\label{fig:bigrama}
\end{figure}

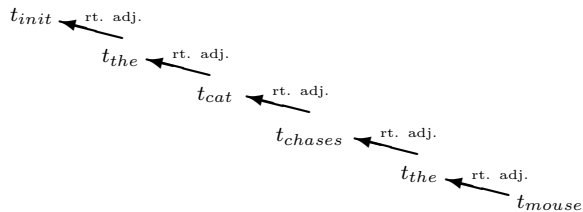
\begin{figure}
\begin{center}
\setlength{\unitlength}{0.00029200in}%
\begingroup\makeatletter\ifx\SetFigFont\undefined
\def\x#1#2#3#4#5#6#7\relax{\def\x{#1#2#3#4#5#6}}%
\expandafter\x\fmtname xxxxxx\relax \def\y{splain}%
\ifx\x\y   
\gdef\SetFigFont#1#2#3{%
  \ifnum #1<17\tiny\else \ifnum #1<20\small\else
  \ifnum #1<24\normalsize\else \ifnum #1<29\large\else
  \ifnum #1<34\Large\else \ifnum #1<41\LARGE\else
     \huge\fi\fi\fi\fi\fi\fi
  \csname #3\endcsname}%
\else
\gdef\SetFigFont#1#2#3{\begingroup
  \count@#1\relax \ifnum 25<\count@\count@25\fi
  \def\x{\endgroup\@setsize\SetFigFont{#2pt}}%
  \expandafter\x
    \csname \romannumeral\the\count@ pt\expandafter\endcsname
    \csname @\romannumeral\the\count@ pt\endcsname
  \csname #3\endcsname}%
\fi
\fi\endgroup
\begin{picture}(11074,5500)(677,-5223)
\put(3006,-517){\makebox(0,0)[lb]{\smash{\SetFigFont{8}{9.6}{rm}cat}}}
\put(4506,-517){\makebox(0,0)[lb]{\smash{\SetFigFont{8}{9.6}{rm}chases}}}
\put(6531,-517){\makebox(0,0)[lb]{\smash{\SetFigFont{8}{9.6}{rm}the}}}
\put(7956,-517){\makebox(0,0)[lb]{\smash{\SetFigFont{8}{9.6}{rm}mouse}}}
\put(1279,-1789){\makebox(0,0)[lb]{\smash{\SetFigFont{8}{9.6}{rm}$t_{init}$}}}
\thicklines
\put(3303,-2089){\vector(-4, 1){1129.412}}
\put(2628,-1789){\makebox(0,0)[lb]{\smash{\SetFigFont{5}{6.0}{rm}rt. adj.}}}
\put(4879,-2764){\vector(-4, 1){1129.412}}
\put(4204,-2464){\makebox(0,0)[lb]{\smash{\SetFigFont{5}{6.0}{rm}rt. adj.}}}
\put(6679,-3439){\vector(-4, 1){1129.412}}
\put(1281,-517){\makebox(0,0)[lb]{\smash{\SetFigFont{8}{9.6}{rm}The}}}
\put(6004,-3139){\makebox(0,0)[lb]{\smash{\SetFigFont{5}{6.0}{rm}rt. adj.}}}
\put(677,-1112){\makebox(0,0)[lb]{\smash{\SetFigFont{8}{9.6}{rm}Corresponding derivation tree:}}}
\put(8629,-4189){\vector(-4, 1){1129.412}}
\put(7954,-3889){\makebox(0,0)[lb]{\smash{\SetFigFont{5}{6.0}{rm}rt. adj.}}}
\put(10279,-4939){\vector(-4, 1){1129.412}}
\put(9604,-4639){\makebox(0,0)[lb]{\smash{\SetFigFont{5}{6.0}{rm}rt. adj.}}}
\put(2928,-2538){\makebox(0,0)[lb]{\smash{\SetFigFont{8}{9.6}{rm}$t_{the}$}}}
\put(4653,-3213){\makebox(0,0)[lb]{\smash{\SetFigFont{8}{9.6}{rm}$t_{cat}$}}}
\put(6078,-3963){\makebox(0,0)[lb]{\smash{\SetFigFont{8}{9.6}{rm}$t_{chases}$}}}
\put(8328,-4638){\makebox(0,0)[lb]{\smash{\SetFigFont{8}{9.6}{rm}$t_{the}$}}}
\put(10428,-5163){\makebox(0,0)[lb]{\smash{\SetFigFont{8}{9.6}{rm}$t_{mouse}$}}}
\put(677, 13){\makebox(0,0)[lb]{\smash{\SetFigFont{8}{9.6}{rm}Example sentence:}}}
\end{picture}
\end{center}
\caption{An example sentence.  Because each tree is right
adjoined to the tree anchored with the neighboring word in the
sentence, the only structure is right branching.}
\label{fig:bigramb}
\end{figure}

The technique used to induce a grammar is a subtractive process.
Starting from a universal grammar (i.e., one that can generate any
string made up of the alphabet set), the parameters are iteratively
refined until the grammar generates, hopefully, all and only the
sentences in the target language, for which the training data provides
an adequate sampling.  In the case of a PCFG, the initial grammar
production rule set contains all possible rules in Chomsky Normal Form
constructed by the nonterminal and terminal symbols.  The initial
parameters associated with each rule are randomly generated subject to
an admissibility constraint.  As long as all the rules have a non-zero
probability, any string has a non-zero chance of being generated.  To
train the grammar, we follow the Inside-Outside re-estimation
algorithm described by \newcite{Lari:90a}.  The Inside-Outside
re-estimation algorithm can also be extended to train PLTIGs.  The
equations calculating the inside and outside probabilities for PLTIGs
can be found in Appendix \ref{app:io}.

As with PCFGs, the initial grammar must be able to generate any
string.  A simple PLTIG that fits the requirement is one that
simulates a bigram model.  It is represented by a tree set that
contains a right auxiliary tree for each lexical item as depicted in
Figure \ref{fig:bigrama}.  Each tree has one adjunction site into which
other right auxiliary trees can adjoin.  The tree set has only one
initial tree, which is anchored by an empty lexical item.  The initial
tree represents the start of the sentence.  Any string can be
constructed by right adjoining the words together in order.  Training
the parameters of this grammar yields the same result as a bigram
model: the parameters reflect close correlations between words that
are frequently seen together, but the model cannot provide any
high-level linguistic structure. (See example in Figure
\ref{fig:bigramb}.)

\begin{figure}[t]
\psfig{figure=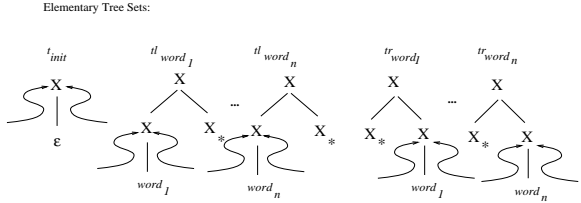,width=3in}
\caption{An LTIG elementary tree set that allow both left and right
adjunctions.}
\label{fig:l1r1a}
\end{figure}

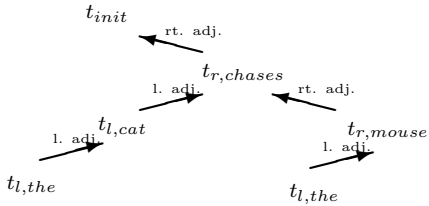
\begin{figure}[h]
\setlength{\unitlength}{0.00029200in}%
\begingroup\makeatletter\ifx\SetFigFont\undefined
\def\x#1#2#3#4#5#6#7\relax{\def\x{#1#2#3#4#5#6}}%
\expandafter\x\fmtname xxxxxx\relax \def\y{splain}%
\ifx\x\y   
\gdef\SetFigFont#1#2#3{%
  \ifnum #1<17\tiny\else \ifnum #1<20\small\else
  \ifnum #1<24\normalsize\else \ifnum #1<29\large\else
  \ifnum #1<34\Large\else \ifnum #1<41\LARGE\else
     \huge\fi\fi\fi\fi\fi\fi
  \csname #3\endcsname}%
\else
\gdef\SetFigFont#1#2#3{\begingroup
  \count@#1\relax \ifnum 25<\count@\count@25\fi
  \def\x{\endgroup\@setsize\SetFigFont{#2pt}}%
  \expandafter\x
    \csname \romannumeral\the\count@ pt\expandafter\endcsname
    \csname @\romannumeral\the\count@ pt\endcsname
  \csname #3\endcsname}%
\fi
\fi\endgroup
\begin{picture}(8294,6029)(301,-5527)
\put(2630,-441){\makebox(0,0)[lb]{\smash{\SetFigFont{8}{9.6}{rm}cat}}}
\put(4130,-441){\makebox(0,0)[lb]{\smash{\SetFigFont{8}{9.6}{rm}chases}}}
\put(6155,-441){\makebox(0,0)[lb]{\smash{\SetFigFont{8}{9.6}{rm}the}}}
\put(7580,-441){\makebox(0,0)[lb]{\smash{\SetFigFont{8}{9.6}{rm}mouse}}}
\thicklines
\put(1352,-4787){\vector( 4, 1){1129.412}}
\put(1577,-4487){\makebox(0,0)[lb]{\smash{\SetFigFont{5}{6.0}{rm}l. adj.}}}
\put(3152,-3887){\vector( 4, 1){1129.412}}
\put(3377,-3587){\makebox(0,0)[lb]{\smash{\SetFigFont{5}{6.0}{rm}l. adj.}}}
\put(4277,-2837){\vector(-4, 1){1129.412}}
\put(3602,-2537){\makebox(0,0)[lb]{\smash{\SetFigFont{5}{6.0}{rm}rt. adj.}}}
\put(905,-441){\makebox(0,0)[lb]{\smash{\SetFigFont{8}{9.6}{rm}The}}}
\put(2177,-2237){\makebox(0,0)[lb]{\smash{\SetFigFont{8}{9.6}{rm}$t_{init}$}}}
\put(5851,-5461){\makebox(0,0)[lb]{\smash{\SetFigFont{8}{9.6}{rm}$t_{l,the}$}}}
\put(6677,-3887){\vector(-4, 1){1129.412}}
\put(6002,-3587){\makebox(0,0)[lb]{\smash{\SetFigFont{5}{6.0}{rm}rt. adj.}}}
\put(6227,-4937){\vector( 4, 1){1129.412}}
\put(6452,-4637){\makebox(0,0)[lb]{\smash{\SetFigFont{5}{6.0}{rm}l. adj.}}}
\put(301,238){\makebox(0,0)[lb]{\smash{\SetFigFont{8}{9.6}{rm}Example sentence:}}}
\put(301,-1336){\makebox(0,0)[lb]{\smash{\SetFigFont{8}{9.6}{rm}Corresponding derivation tree:}}}
\put(2401,-4261){\makebox(0,0)[lb]{\smash{\SetFigFont{8}{9.6}{rm}$t_{l,cat}$}}}
\put(751,-5311){\makebox(0,0)[lb]{\smash{\SetFigFont{8}{9.6}{rm}$t_{l,the}$}}}
\put(4276,-3286){\makebox(0,0)[lb]{\smash{\SetFigFont{8}{9.6}{rm}$t_{r,chases}$}}}
\put(6901,-4336){\makebox(0,0)[lb]{\smash{\SetFigFont{8}{9.6}{rm}$t_{r,mouse}$}}}
\end{picture}
\caption{ With both left and right adjunctions possible, the sentences
can be parsed in a more linguistically plausible way}
\label{fig:l1r1b}
\end{figure}

To generate non-linear structures, we need to allow adjunction in both
left and right directions.  The expanded LTIG tree set includes a left
auxiliary tree representation as well as right for each lexical item.
Moreover, we must modify the topology of the auxiliary trees so that
adjunction in both directions can occur.  We insert an intermediary
node between the root and the lexical word.  At this internal node, at
most one adjunction of each direction may take place.  The
introduction of this node is necessary because the definition of the
formalism disallows right adjunction into the root node of a left
auxiliary tree and vice versa.  For the sake of uniformity, we shall
disallow adjunction into the root nodes of the auxiliary trees from
now on.  Figure
\ref{fig:l1r1a} shows an LTIG that allows at most one left and one
right adjunction for each elementary tree.  This enhanced LTIG can
produce hierarchical structures that the bigram model could not (See
Figure \ref{fig:l1r1b}.)

It is, however, still too limiting to allow only one adjunction from
each direction.  Many words often require more than one modifier.  For
example, a transitive verb such as ``give'' takes at least two
adjunctions: a direct object noun phrase, an indirect object noun
phrase, and possibly other adverbial modifiers.  To create more
adjunction sites for each word, we introduce yet more intermediary
nodes between the root and the lexical word.  Our empirical studies
show that each lexicalized auxiliary tree requires at least 3
adjunction sites to parse all the sentences in the corpora.  Figure
\ref{fig:LnRm}(a) and (b) show two examples of auxiliary trees with 3
adjunction sites.  The number of parameters in a PLTIG is dependent on
the number of adjunction sites just as the size of a PCFG is dependent
on the number of nonterminals.  For a language with $V$ vocabulary
items, the number of parameters for the type of PLTIGs used in this
paper is $2(V+1) + 2V(K)(V+1)$, where $K$ is the number of adjunction
sites per tree.  The first term of the equation is the number of
parameters contributed by the initial tree, which always has two
adjunction sites in our experiments.  The second term is the
contribution from the auxiliary trees.  There are $2V$ auxiliary
trees, each tree has $K$ adjunction sites; and $V+1$ parameters
describe the distribution of adjunction at each site.  The number of
parameters of a PCFG with M nonterminals is $M^3 + MV$.  For the
experiments, we try to choose values of $K$ and $M$ for the PLTIGs and
PCFGs such that
\[
	2(V+1) + 2V(K)(V+1) \approx  M^3+MV
\]

\subsection{ATIS}
\label{atis}

\begin{figure}[t]
\begin{center}
\begin{tabular}{c|c|c}
  \mbox{\psfig{figure=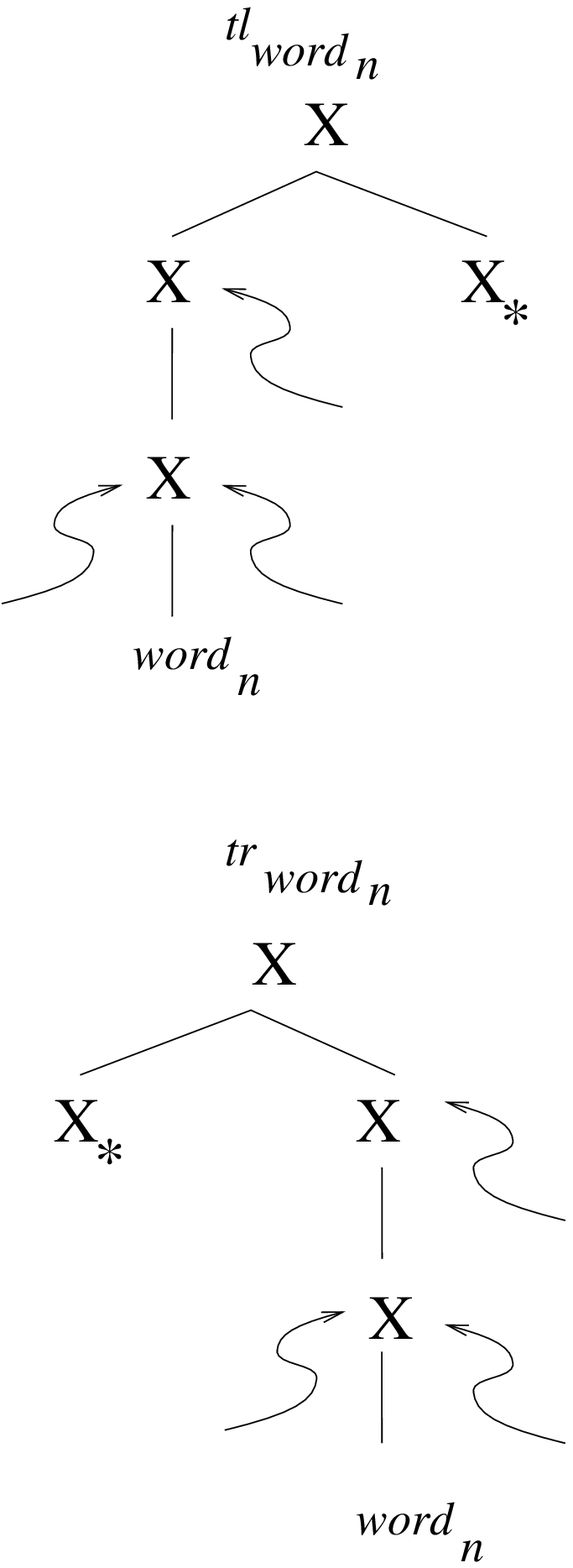,height=2.5in}} 
& \mbox{\psfig{figure=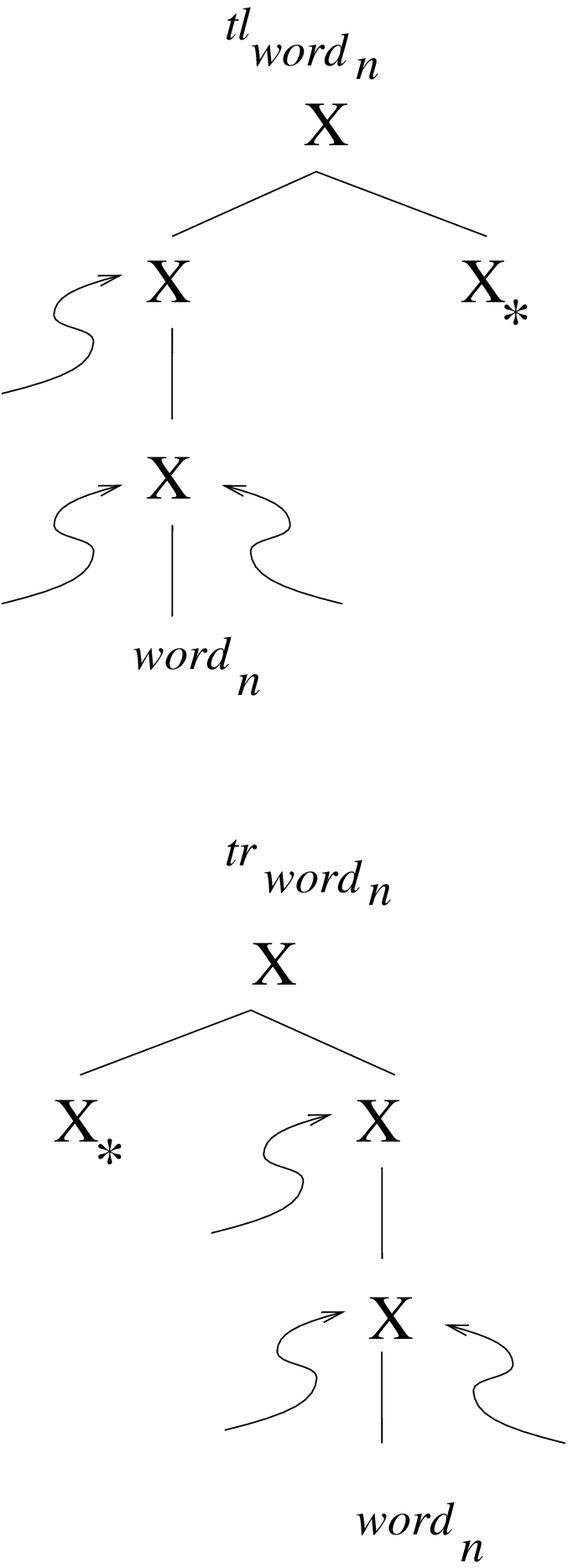,height=2.5in}} 
& \mbox{\psfig{figure=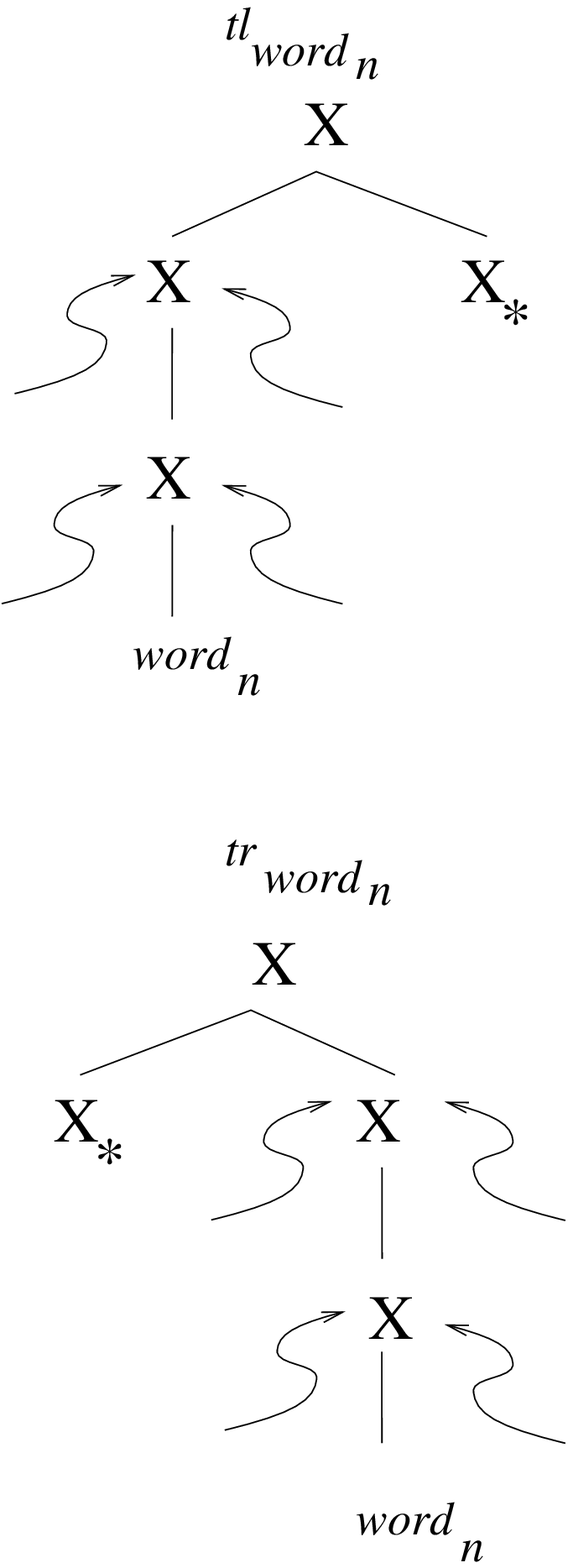,height=2.5in}} \\
(a) & (b) & (c)
\end{tabular}
\end{center}
\caption{Prototypical auxiliary trees for three PLTIGs: (a) L1R2, (b)
L2R1, and (c) L2R2.}
\label{fig:LnRm}
\end{figure}

To reproduce the results of PCFGs reported by Pereira and Schabes, we
use the ATIS corpus for our first experiment.  This corpus contains
577 sentences with 32 part-of-speech tags.  To ensure statistical
significance, we generate ten random train-test splits on the corpus.
Each set randomly partitions the corpus into three sections according
to the following distribution: 80\% training, 10\% held-out, and 10\%
testing.  This gives us, on average, 406 training sentences, 83
testing sentences, and 88 sentences for held-out testing.  The results
reported here are the averages of ten runs.

We have trained three types of PLTIGs, varying the number of left and
right adjunction sites.  The L2R1 version has two left adjunction
sites and one right adjunction site; L1R2 has one left adjunction site
and two right adjunction sites; L2R2 has two of each.  The
prototypical auxiliary trees for these three grammars are shown in
Figure \ref{fig:LnRm}.  At the end of every training iteration, the
updated grammars are used to parse sentences in the held-out test sets
$D$, and the new language modeling scores (by measuring the
cross-entropy estimates $\hat{H}(D, L2R1)$, $\hat{H}(D, L1R2)$, and
$\hat{H}(D, L2R2)$) are calculated.  The rate of improvement of the
language modeling scores determines convergence.  The PLTIGs are
compared with two PCFGs: one with 15-nonterminals, as Pereira and
Schabes have done, and one with 20-nonterminals, which has comparable
number of parameters to L2R2, the larger PLTIG.

\begin{figure}[t]
\psfig{figure=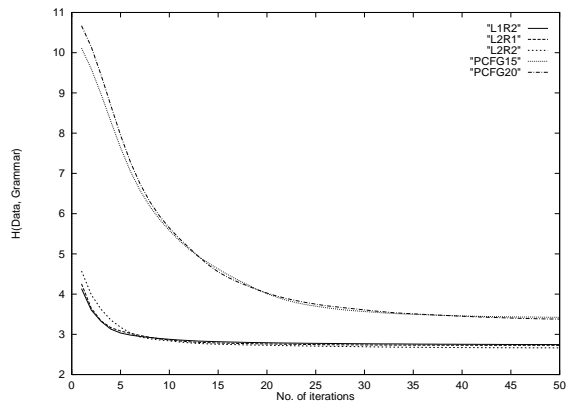,width=3.0in}
\caption{Average convergence rates of the training process for 3 PLTIGs and 2
PCFGs.}
\label{fig:devs}
\end{figure}

In Figure \ref{fig:devs} we plot the average iterative improvements of
the training process for each grammar.  All training processes of the
PLTIGs converge much faster (both in numbers of iterations and in real
time) than those of the PCFGs, even when the PCFG has fewer parameters
to estimate, as shown in Table \ref{tab:atis}.  From Figure
\ref{fig:devs}, we see that both PCFGs take many more iterations to
converge and that the cross-entropy value they converge on is much
higher than the PLTIGs.

During the testing phase, the trained grammars are used to produce
bracketed constituents on unmarked sentences from the testing sets
$T$.  We use the crossing bracket metric to evaluate the parsing
quality of each grammar.  We also measure the cross-entropy estimates
$\hat{H}(T,L2R1)$, $\hat{H}(T,L1R2)$,$\hat{H}(T,L2R2)$,
$\hat{H}(T,PCFG_{15})$, and $\hat{H}(T,PCFG_{20})$ to determine the
quality of the language model.  For a baseline comparison, we consider
bigram and trigram models with simple right branching bracketing
heuristics.  Our findings are summarized in Table \ref{tab:atis}.

\begin{table*} 
\begin{tabular} 
{| p{1.9in} || c |c|c|c|c|c|} \hline & Bigram/Trigram & PCFG 15 & PCFG 20
                         & L1R2 & L2R1 & L2R2 \\ \hline \hline Number of
                         parameters & 1088 / 34880 & 3855 & 8640 & 6402 &
                         6402 & 8514 \\ \hline Iterations to convergence
                         & -- & 45 & 45 & 19 & 17 & 24 \\ \hline
                         Real-time convergence (min)& -- & 62 & 142 & 8 &
                         7 & 14 \\ \hline $\hat{H}(T,Grammar)$ & 2.88 /
                         2.71 & 3.81 & 3.42 & 2.87 & 2.85 & 2.78 \\
                         \hline Crossing bracket (on $T$) & 66.78 & 93.46
                         & 93.41 & 93.07 & 93.28 & 94.51 \\ \hline
\end{tabular}
\caption{Summary results for ATIS.  The machine used to measure real-time
is an HP 9000/859. }
\label{tab:atis}
\end{table*}

The three types of PLTIGs generate roughly the same number of
bracketed constituent errors as that of the trained PCFGs, but they
achieve a much lower entropy score.  While the average entropy value
of the trigram model is the lowest, there is no statistical
significance between it and any of the three PLTIGs.  The relative
statistical significance between the various types of models is
presented in Table \ref{tab:sig}.  In any case, the slight language
modeling advantage of the trigram model is offset by its inability to
handle parsing.

\begin{table}
\begin{tabular}
{ l ||ccc}  
PLTIGs   &  better   &	        &          \\ 
bigram 	 &  better   &	 --     &          \\ 
trigram	 &  better   &	 --     &  better  \\ \hline \hline
         &   PCFGs   &  PLTIGs  &  bigram  \\ \hline
\end{tabular}
\caption{Summary of pair-wise t-test for all grammars.  If ``better''
appears at cell ($i$,$j$), then the model in row $i$ has an entropy value
lower than that of the model in column $j$ in a statistically significant
way.  The symbol ``--'' denotes that the difference of scores between the
models bears no statistical significance. }
\label{tab:sig}
\end{table}

Our ATIS results agree with the findings of Pereira and Schabes that
concluded that the performances of the PCFGs do not seem to depend
heavily on the number of parameters once a certain threshold is
crossed.  Even though $PCFG_{20}$ has about as many number of
parameters as the larger PLTIG (L2R2), its language modeling score is
still significantly worse than that of any of the PLTIGs.

\subsection{WSJ}
\label{wsj}


Because the sentences in ATIS are short with simple and similar
structures, the difference in performance between the formalisms may not
be as apparent.  For the second experiment, we use the Wall Street
Journal (WSJ) corpus, whose sentences are longer and have more varied and
complex structures.  We use sections 02 to 09 of the WSJ corpus for
training, section 00 for held-out data $D$, and section 23 for test $T$.
We consider sentences of length 40 or less.  There are 13242 training
sentences, 1780 sentences for the held-out data, and 2245 sentences in
the test.  The vocabulary set consists of the 48 part-of-speech tags.  We
compare three variants of PCFGs (15 nonterminals, 20 nonterminals, and 23
nonterminals) with three variants of PLTIGs (L1R2, L2R1, L2R2).  A PCFG
with 23 nonterminals is included because its size approximates that of
the two smaller PLTIGs.  We did not generate random train-test splits for
the WSJ corpus because it is large enough to provide adequate sampling.
Table \ref{tab:wsj} presents our findings.
\begin{table*}
\begin{tabular} 
{| p{1in} ||c|c|c|c|c|c|c|} \hline & Bigram/Trigram & PCFG 15 & PCFG 20 &
                        PCFG 23 & L1R2 & L2R1 & L2R2 \\ \hline \hline
                        Number of parameters & 2400 / 115296 & 4095 &
                        8960 & 13271 & 14210 & 14210 & 18914 \\ \hline
                        Iterations to convergence & -- & 80 & 60 & 70 &
                        28 & 30 & 28 \\ \hline Real-time convergence
                        (hr)& -- & 143 & 252 & 511 & 38 & 41 & 60 \\
                        \hline
$\hat{H}(T,Grammar)$ & 3.39/3.20 & 4.31 & 4.27 & 4.13 & 3.58 & 3.56 &
3.59 \\ \hline Crossing bracket (T) & 49.44 & 56.41 & 78.82 & 79.30 &
80.08 & 82.43 & 80.832 \\ \hline
\end{tabular}
\caption{Summary results for WSJ}
\label{tab:wsj}
\end{table*}
From Table \ref{tab:wsj}, we see several similarities to the results
from the ATIS corpus.  All three variants of the PLTIG formalism have
converged at a faster rate and have far better language modeling
scores than any of the PCFGs.  Differing from the previous experiment,
the PLTIGs produce slightly better crossing bracket rates than the
PCFGs on the more complex WSJ corpus.  At least 20 nonterminals are
needed for a PCFG to perform in league with the PLTIGs.  Although the
PCFGs have fewer parameters, the rate seems to be indifferent to the
size of the grammars after a threshold has been reached.  While upping
the number of nonterminal symbols from 15 to 20 led to a 22.4\% gain,
the improvement from $PCFG_{20}$ to $PCFG_{23}$ is only 0.5\%.
Similarly for PLTIGs, L2R2 performs worse than L2R1 even though it has
more parameters.  The baseline comparison for this experiment results
in more extreme outcomes.  The right branching heuristic receives a
crossing bracket rate of 49.44\%, worse than even that of $PCFG_{15}$.
However, the $N$-gram models have better cross-entropy measurements
than PCFGs and PLTIGs; bigram has a score of 3.39 bits per word, and
trigram has a score of 3.20 bits per word.  Because the lexical
relationship modeled by the PLTIGs presented in this paper is limited
to those between two words, their scores are close to that of the
bigram model.

\section{Conclusion and Future Work}

In this paper, we have presented the results of two empirical experiments
using Probabilistic Lexicalized Tree Insertion Grammars.  Comparing
PLTIGs with PCFGs and $N$-grams, our studies show that a lexicalized tree
representation drastically improves the quality of language modeling of a
context-free grammar to the level of $N$-grams without degrading the
parsing accuracy.  In the future, we hope to continue to improve on the
quality of parsing and language modeling by making more use of the
lexical information.  For example, currently, the initial untrained
PLTIGs consist of elementary trees that have uniform configurations
(i.e., every auxiliary tree has the same number of adjunction sites) to
mirror the CNF representation of PCFGs.  We hypothesize that a grammar
consisting of a set of elementary trees whose number of adjunction sites
depend on their lexical anchors would make a closer approximation to the
``true'' grammar.  We also hope to apply PLTIGs to natural language tasks
that may benefit from a good language model, such as speech recognition,
machine translation, message understanding, and keyword and topic
spotting.

\begin{appendix}

\section{Parameters of PLTIG}
\label{app:param}

Each elementary tree of a Probabilistic Tree Insertion Grammar,
denoted as $\rho$, has the following parameters:

\begin{description}
\item $P_I(\rho)$: the probability that tree $\rho$ is the start of a
derivation (i.e., tree $\rho$ does not adjoin or substitute into other
trees).  If $\rho$ is an auxiliary tree, $P_I(\rho)=0$.  The grammars
used in our experiments have exactly one empty initial tree with
$P_I(\rho_\in)=1$.
\end{description}
The parameters for adjunction and substitution are associated with
each node of an elementary tree, denoted as $\eta$.
\begin{description}
\item $P_L(\eta, \rho_L)$: the probability of adjunction between left
auxiliary tree $\rho_L$ and node $\eta$
\item $P_{NL}(\eta)$: the probability that no tree left adjoins into
node $\eta$ such that
\[
	P_{NL}(\eta) + \sum\limits_{\rho_L}{P_L(\eta,\rho_L)} = 1
\]
\item $P_R(\eta, \rho_R)$: the probability of adjunction between right
auxiliary tree $\rho_R$ and node $\eta$
\item $P_{NR}(\eta)$: the probability that no tree right adjoins into node
$\eta$ such that
\[
	P_{NR}(\eta) + \sum\limits_{\rho_R}{P_R(\eta,\rho_R)} = 1
\]
\item $P_S(\eta, \rho_S)$: the probability that an initial tree
$\rho_S$ can substitute into node $\eta$.  The grammars we used for
our experiments have no substituion nodes, so this parameter is not
used.
\end{description}

\section{Inside-Outside Probabilities}
\label{app:io}

Let ${\bf O} = O_1,O_2,\ldots,O_T$ be the observed sequence we wish to
parse with a PLTIG.  To estimate the likelihood of observing this
sequence in the grammar and to maximize the parameters of the grammar
to reflect the observations, we compute the inside and outside
probabilities.

\subsection{Inside Probabilities}

The inside probability of a node $\eta$ between positions $s$ and $t$,
is the probability that node $\eta$ can generate the partial
observations between $s$ and $t$ (i.e., $O_{s+1},\ldots,O_t$).  This
probability is denoted as $\RHnorm{e}(s,t,\eta)$.  We calculate $\RHnorm{e}$
recursively in a bottom-up manner.  The base cases are when a node is
an empty node or a foot node, which does not cover anything; and when
a node covers a single lexical item (e.g., $O_{s+1}$).
\[
\begin{array}{rcl}
	\RHnorm{e}(s,s,\eta) & = & \left\{ \begin{array}{l@{\quad:\quad}l}
		\SC 1 & \SC{Foot(\eta) \hspace{1ex}\mbox{or}\hspace{1ex}
		Label(\eta) = \epsilon} \\ \SC 0 & \SC{\mbox{otherwise} }
		\end{array} \right. \\ \RHnorm{e}(s,s+1,\eta) & = & \left\{
		\begin{array}{l@{\quad:\quad}l} \SC 1 & \SC{Label(\eta) =
		O_{s+1}} \\ \SC 0 & \SC{\mbox{otherwise} } \end{array}
		\right. \\
\end{array}
\]

We now show the general case of computing the inside probability for a
node $\eta$ generating the sub-sequence of observations between
positions $s$ and $t$.  Following the model outlined in
\newcite{Schabes:93b}, we enforce the restriction that a node cannot
have more than one left or right adjunction.  More specifically, there
are four ways that node $\eta$ can generate the sub-sequence between
positions $s$ and $t$:
\begin{enumerate}
\item $e(s,t,\eta, \emptyset)$: the probability that $\eta$ covers
$O_{s+1},\ldots, O_t$ without any adjunction at $\eta$. 
\item $e(s,t,\eta, L)$: exactly one left adjunction at $\eta$. 
\item $e(s,t,\eta, R)$: exactly one right adjunctions at $\eta$. 
\item $e(s,t,\eta, LR)$ a simultaneous left and right adjunction at $\eta$.  
\end{enumerate}
The final value of $\RHnorm{e}(s,t,\eta)$ is the normalized sum of the
four parts.
\[
\begin{array}{rcl}
	\RHnorm{e}(s,t,\eta) & = &
                          \SC{P_{NL}(\eta)P_{NR}(\eta)e(s,t,\eta,\emptyset)
                          }\\ & + & \SC{ P_{NR}(\eta)e(s,t,\eta,L) }\\
                          & + & \SC{ P_{NL}(\eta)e(s,t,\eta,R) }\\ & +
                          & \SC{e(s,t,\eta,LR)/2 }\\
\end{array}
\]

For a node $\eta$ to cover the substring between positions $s$ and $t$
without any adjunction, it must be the case that its children jointly
generate $O_{s+1},\ldots, O_t$.\footnote{Or, if $\eta$ were a
substitution node, then there must be a tree that generate the
substring and can be substituted into $\eta$.  We did not include the
equations for this case because our grammars have no substitution
nodes.}  If node $\eta$ has only one child $\eta_1$, then
\[
	e(s,t,\eta,\emptyset) = \RHnorm{e}(s,t,\eta_1)
\]
If node $\eta$ has two children such that $\eta_1$ is the left child
and $\eta_2$ is the right child, then
\[
	e(s,t,\eta,\emptyset) =
\sum\limits_{r=s}^t{\RHnorm{e}(s,r,\eta_1)\RHnorm{e}(r,t,\eta_2)}
\]
Next, we consider the case when $\eta$ generates $O_{s+1},\ldots, O_t$
by letting a left auxiliary tree adjoin into it.  The auxiliary tree
generates the front of the substring, $O_{s+1},\ldots, O_r$, and
$\eta$ generates the rest of the substring, $O_r+1,\ldots, O_t$,
without adjunctions.  The breaking position $r$ can be anywhere
between $s$ and $t$.
\[
\begin{array}{rcl}
	e(s,t,\eta,L) & = & \sum\limits_{\rho_L}{\sum\limits_{r=s+1}^t{
								\left( \begin{array}{l} \SC
								{\RHnorm{e}(s,r,\rho_L)} \\ \SC
								{\times e(r,t,\eta,\emptyset)} \\ \SC
								{\times P_L(\eta,\rho_L)} \end{array}
								\right) }}\\
\end{array}
\]
Similary, $e(s,t,\eta,R)$ represents the case when a right auxiliary
tree, $\rho_R$, is adjoined into node $\eta$ to generate
$O_{s+1},\ldots, O_t$.
\[
\begin{array}{rcl}
	e(s,t,\eta,R) & = & \sum\limits_{\rho_R}{\sum\limits_{r=s}^{t-1}{
								\left( \begin{array}{l} \SC
								{\RHnorm{e}(s,r,\rho_R)} \\ \SC
								{\times e(r,t,\eta,\emptyset)} \\ \SC
								{\times P_R(\eta,\rho_R)} \end{array}
								\right) }}\\
\end{array}
\]

Finally, if both a left auxliary tree and a right auxiliary tree
adjoin into node $\eta$ simultaneously to generate $O_{s+1},\ldots,
O_t$, then we have to consider two breaking positions $r1$ and $r2$.
The left auxiliary tree, $\rho_L$, generates $O_{s+1},\ldots,
O_{r1}$; the node $\eta$ generates $O_{r1+1},\ldots, O_{r2}$ (without
adjunction); and the right auxiliary tree, $\rho_R$, generates the
remaining observations $O_{r2+1},\ldots, O_{t}$.  
\[
\begin{array}{rcl}
	e(s,t,\eta,LR) & = &
    \sum\limits_{\rho_L}{\sum\limits_{\rho_R}{\sum\limits_{r_1=s+1}^{t-1}
{\sum\limits_{r_2=r_1}^{t-1}{\left(\begin{array}{l}
    \SC{\RHnorm{e}(s,r_1,\rho_L) }\\ \SC{\times
    e(r_1,r_2,\eta,\emptyset)} \\ \SC{\times \RHnorm{e}(r_2,t,\rho_R)
    }\\ \SC{\times P_R(\eta,\rho_R)} \\ \SC{\times P_L(\eta,\rho_L)}
    \end{array}\right) }}}}
\end{array}
\]

\subsection{Outside Probabilities}

The outside probability of a node $\eta$ between positions $s$ and
$t$, denoted as $\RHnorm{f}(s,t,\eta)$, is the probability that the
derived tree will generate $\eta$ and the two partial observations
{\it outside of} $s$ and $t$ (i.e., the two sub-sequences $O_1,
\ldots, O_s$ and $O_{t+1}, \ldots, O_{T}$).  The outside probabilities
complement the inside probabilities: the product of the matching
inside and outside probabilities is the total probability of the
observation sequence being generated by the grammar.  Similar to the
constructs of the inside probabilities, we define four types of
outside probabilities:
\begin{enumerate}
\item $f(s,t,\eta,\emptyset)$: the probability that $\eta$ is
generated without having any tree adjoining into it.
\item $f(s,t,\eta,L)$: $\eta$ is generated and a left auxiliary tree
has adjoined into it.  Moreover, the auxiliary tree does not cover any
part of the substring between $s$ and $t$.
\item $f(s,t,\eta,R)$: $\eta$ is generated and a right auxiliary tree
has adjoined into it.  Moreover, the auxiliary tree does not cover any
part of the substring between $s$ and $t$.
\item $f(s,t,\eta,LR)$: $\eta$ is generated and a left auxiliary tree
and a right auxiliary tree have simultaneously adjoined into it.
Neither auxiliary tree can cover any part of the substring between $s$
and $t$.
\end{enumerate}
Finally, $\RHnorm{f}(s,t,\eta)$ is the normalized sum of its four parts.
\[
\begin{array}{rcl}
	\RHnorm{f}(s,t,\eta) & = &
				\SC{P_{NL}(\eta)P_{NR}(\eta)f(s,t,\eta,\emptyset) }\\ & +
				& \SC{ P_{NR}(\eta)f(s,t,\eta,L) }\\ & + & \SC{
				P_{NL}(\eta)f(s,t,\eta,R) }\\ & + & \SC{ f(s,t,\eta,LR)/2
				}\\
\end{array}
\]

The outside probabilities are computed recursively in a top-down
manner.  The base case is the probability that the root node $\eta$ of
an initial tree $\rho$ is generated.  This is equal to the probability
of the initial tree being the start of the derivation.
\[
f(0,T,\eta, \emptyset) = \left\{ \begin{array}{l@{\quad:\quad}l}
	 \SC{P_I(\rho)} & \SC{IsRoot(\eta, \rho)}  \\
	 \SC 0	       & \SC{ \mbox{otherwise}}
	\end{array} \right.
\]

To compute the outside probability of a node without any auxiliary
trees adjoining into it, we consider five \footnote{For the work
presented here, we do not consider the sixth case in which the node is
the root of an initial tree that might substitute into a substitution
node.} different tree configurations.
\begin{itemize}
\item node $\eta$ is the only child of its parent node, $\eta_0$.
Because no adjunction took place at $\eta$, its outside probability is
the normalized outside probability of its parent node.
\[
	f(s,t,\eta,\emptyset) = \RHnorm{f}(s,t,\eta_0)
\]
\item node $\eta$ is the left child of node $\eta_0$ and $\eta$ has
a sibling $\eta_2$, the right child of $\eta_0$.  The outside
probability of $\eta$ between positions $s$ and $t$ is the product of
the outside probability of its parent node between positions $s$ and
$r$, where $t<r \le T$ and the normalized inside probability of its
sibling node $\eta_2$ deriving the substring between $t$ and $r$.
\[
	f(s,t,\eta,\emptyset) =
\sum\limits_{r=t}^T{\RHnorm{f}(s,r,\eta_0)\RHnorm{e}(t,r,\eta_2)}
\]
\item node $\eta$ is the right child of node $\eta_0$ and $\eta$ has a
sibling $\eta_1$, the left child of $\eta_0$.
\[
	f(s,t,\eta,\emptyset) =
\sum\limits_{r=0}^s{\RHnorm{f}(r,t,\eta_0)\RHnorm{e}(r,s,\eta_1)}
\]
\item node $\eta$ is the root node of a left auxiliary tree $\rho_L$
that left adjoins into a node $\eta_0$.  Suppose that node $\eta_0$
derives the substring between positions $t$ and $r$, where $t < r \le
T$.  Then the outside probability of $\eta$ between $s$ and $t$ is the
product of the outside probability of $\eta_0$ between $s$ and $r$ and
the inside probability of $\eta_0$ deriving the observations between
$t$ and $r$ without left adjunction.  In order for $\rho_L$ to left
adjoin into $\eta_0$, $\eta_0$ must not have previously left adjoined
with any tree.  Therefore, the inside probability of $\eta_0$ between
$t$ and $r$ cannot include $e(t,r,\eta_0,L)$ or $e(t,r,\eta_0,LR)$.
\[
    f(s,t,\eta,\emptyset) = \sum\limits_{\eta_0} {\sum\limits_{r=t}^T{
		\left( \begin{array}{l} \SC{
		P_L(\eta_0,\rho_L)f(s,r,\eta_0,\emptyset)} \\ \SC{\times} \left[
		\begin{array}{l} \SC{e(t,r,\eta_0,\emptyset)P_{NR}(\eta_0)} \\
		\SC{+ e(t,r,\eta_0,R)/2} \end{array}\right] \end{array} \right)}}
\]
\item node $\eta$ is the root node of a right auxiliary tree $\rho_R$
that right adjoins into a node $\eta_0$.  Suppose node $\eta_0$
generates the partial string between position $r$ and $s$, where $0
\le r < s$; then the outside probability of $\eta$ between $s$ and $t$
is the product of the outside probability of $\eta_0$ between $r$ and
$t$ and the inside probability of $\eta_0$ between $r$ and $s$ without
any right adjunction.
\[
    f(s,t,\eta,\emptyset) = \sum\limits_{\eta_0}{\sum\limits_{r=0}^s{
		\left( \begin{array}{l}
		\SC{P_R(\eta_0,\rho_R)f(r,t,\eta_0,\emptyset)} \\ \SC{\times}
		\left[ \begin{array}{l}
		\SC{e(r,s,\eta_0,\emptyset)P_{NL}(\eta_0)} \\ \SC{+
		e(r,s,\eta_0,L)/2} \end{array}\right] \end{array} \right)}}
\]
\end{itemize}

The remaining three types of outside probabilities are the cases in which
auxiliary trees are adjoined into node $\eta$.  First, we consider the
case of left adjunction.  Let tree $\rho_L$ be an auxiliary tree that
is to be adjoined into $\eta$.  $\rho_L$ must derive the partial observation
immediately before position $s$ (i.e., $O_r, \ldots, O_s$, where
$0 \le r < s$).  
\[
	f(s,t,\eta,L) = \sum\limits_{\rho_L}{\sum\limits_{r=0}^s{ \left(
							\begin{array}{l} \SC{ \RHnorm{e}(r,s,\rho_L)
							} \\ \SC{ \times f(r,t,\eta,\emptyset)} \\
							\SC{ \times P_L(\eta,\rho_L) } \end{array}
							\right) }}\\
\]
Similarly, if a right auxiliary tree, $\rho_R$, is to be adjoined into
node $\eta$, then it must derive the partial observation immediately
after position $t$ (i.e., $O_{t+1}, \ldots, O_r$, where $t < r \le
T$).
\[
	f(s,t,\eta,R) = \sum\limits_{\rho_R}{\sum\limits_{r=t}^T{ \left(
							\begin{array}{l} \SC{ \RHnorm{e}(t,r,\rho_R)}
							\\ \SC{ \times f(s,r,\eta,\emptyset)} \\ \SC{
							\times P_R(\eta,\rho_R)} \end{array} \right)
							}}\\
\]
Finally, in the case of simultaneous adjunction, both a left auxiliary
tree $\rho_L$ and a right auxiliary tree $\rho_R$ are adjoined into
node $\eta$ such that $\rho_L$ covers a partial string from position $r_1$
to $s$ and $\rho_R$ covers a partial string from position $t$ to
$r_2$, where $0 \le r_1< s$ and $t < r_2 \le T$.
\[
	f(s,t,\eta,LR) =
	\sum\limits_{\rho_L}{\sum\limits_{
\rho_R}{\sum\limits_{r_1=0}^s{\sum\limits_{r_2=t}^T{\left(\begin{array}{l}
	\SC{\RHnorm{e}(r_1,s,\rho_L)} \\ \SC{\times \RHnorm{e}(t,r_2,\rho_R)}
	\\ \SC{\times f(r_1,r_2,\eta,\emptyset)} \\ \SC{\times
	P_R(\eta,\rho_R)} \\ \SC{\times P_L(\eta,\rho_L)} \end{array}\right)
	}}}}
\]

\end{appendix}

\end{document}